%% file: main.tex
\documentclass[10pt,twocolumn,letterpaper]{article}

\usepackage[pagenumbers]{cvpr}
\input{preamble}

\definecolor{cvprblue}{rgb}{0.21,0.49,0.74}
\usepackage[pagebackref,breaklinks,colorlinks,allcolors=cvprblue]{hyperref}

\title{PhasePlan: Ordered Future-Phase Planning for Robot Brain Models}

\author{
Xiaoyu Yang$^{*}$ \quad
Yafei Zhang$^{*}$ \quad
Wensheng Li \\
Qing Zhan \quad
Nan Wu$^{\dagger}$\\
\vspace{0.5em}
\small $^{*}$Equal contribution. \qquad
$^{\dagger}$Corresponding author: \texttt{across2026@163.com}
}

\begin{document}
\maketitle

\input{sec/0_abstract}
\input{sec/1_introduction}
\input{sec/1_related_work}
\input{sec/2_method}
\input{sec/3_experiments}
\input{sec/4_results}
\input{sec/5_conclusion}

{\small
\bibliographystyle{ieeenat_fullname}
\bibliography{references}
}

\input{sec/6_supplement}

\end{document}

%% file: preamble.tex
\usepackage{amsmath,amsfonts,amssymb}
\usepackage{booktabs}
\usepackage{makecell}
\usepackage{microtype}
\usepackage{xspace}

\newcommand{\method}{PhasePlan\xspace}
\newcommand{\R}{\mathbb{R}}

%% file: sec/0_abstract.tex
\begin{abstract}
Robot brain models integrate vision, language, and robot state to generate actions for complex manipulation tasks.
Most predict fixed-length action chunks that may span multiple task phases.
This can obscure phase transitions and favor frequent action patterns, compromising action timing in dynamic environments.
We propose \method, an ordered future-phase planning method for robot brain models.
From current multimodal observations, it predicts the task phase at each future action position.
The resulting planning representations condition the corresponding actions, preserving temporal alignment between task progress and action generation.
Training first learns the planner, then freezes it during action-model adaptation to maintain stable phase representations.
We instantiate \method on pretrained $\pi_{0.5}$ and AcrossWAM1.0 robot brain models.
Detailed quantitative evaluation uses the $\pi_{0.5}$ implementation.
On conveyor-belt manipulation, \method reduces offline joint-action error by approximately 22.5\% relative to the original $\pi_{0.5}$ model.
It also improves phase-transition modeling and cross-phase action prediction.
These results demonstrate the value of ordered future-phase planning for continuous action generation.
\end{abstract}

%% file: sec/1_introduction.tex
\section{Introduction}
\label{sec:introduction}

Robot brain models connect multimodal perception, task planning, and motion prediction to support robotic manipulation.
Vision-language-action (VLA) policies translate these capabilities into executable robot actions~\cite{ji2025robobrain,black2025pi05}.
Pretrained policies make downstream adaptation attractive when task-specific demonstrations are costly to collect.

Downstream adaptation must account for temporal regularity and phase imbalance in demonstrations.
In our conveyor task, the robot waits for an actionable red package.
It then grasps the package and transports it to a basket.
After release, the robot returns to its initial pose.
Waiting and transport take much longer than release and phase transitions.
Action optimization favors frequent motion patterns, while reliable execution also requires precise transition timing.
Action-chunk prediction makes transition timing particularly important.
A single output chunk can cross boundaries such as \textsc{Wait}$\rightarrow$\textsc{Act} or \textsc{Act}$\rightarrow$\textsc{Release}.
Policies with similar average action errors can therefore exhibit different transition timing and closed-loop behavior.

Position-aligned task-progress representations can capture phase order and transition locations within an action chunk.
For example, \textsc{Wait}, \textsc{Wait}, \textsc{Act}, \textsc{Act} and \textsc{Act}, \textsc{Act}, \textsc{Wait}, \textsc{Wait} have identical phase counts.
The two sequences require different actions at each future position.
This distinction motivates a position-aligned representation of task progress within action chunks.

We propose \method, an ordered future-phase planning framework for robot brain models.
We instantiate it on a pretrained $\pi_{0.5}$ VLA policy.
The planner receives head-view RGB, wrist-view RGB, proprioceptive state, and a language instruction.
Independent queries predict a phase distribution for each of the $H=16$ future positions.
Each distribution and its contextual planning representation form a conditioning feature.
The first feature guides the first future action, and the $H$-th feature guides the $H$-th action.
The ordered structure of the predicted plan is thereby carried into action generation.

Training is divided into two stages.
First, the planner learns to predict a complete ordered future phase sequence from the current observation.
The planner is then frozen, and we adapt the action expert module and the plan-conditioning path.
This separation allows the action model to use the learned phase-transition structure.
At deployment, the planner remains an active internal component of the policy.

Experiments on conveyor-belt manipulation validate both the ordered planner and the position-aligned conditioning mechanism.
The independent-query planner achieves about 92.5\% per-position future-phase accuracy and about 89\% phase-transition-type accuracy.
Removing or shuffling the RGB inputs clearly reduces macro-averaged F1.
This indicates that the predicted plan uses the current visual observation.
With a frozen planner and per-position conditioning, offline joint-action error drops by about 22.5\% relative to the original $\pi_{0.5}$.
The reduction relative to the future-skill-histogram baseline is about 12.7\%.

Our contributions are as follows:
\begin{itemize}
  \item \textbf{Ordered future-phase planning.} We use $H$ position-corresponding phase predictions to represent task progress inside an action chunk. This preserves both phase categories and transition positions.
  \item \textbf{Per-position plan conditioning.} Each planning representation and its phase distribution condition the corresponding future action token.
  \item \textbf{Two-stage planning and action adaptation.} The planner is trained first and then frozen during action adaptation. This maintains phase-boundary quality while improving action prediction.
\end{itemize}

%% file: sec/1_related_work.tex
\section{Related Work}
\label{sec:related_work}

\subsection{Robot Brain Models and VLA Policies}

Robot brain models connect multimodal perception, task planning, and motion prediction.
RoboBrain treats these abilities as complementary robot capabilities~\cite{ji2025robobrain}.
Vision-language-action (VLA) policies provide an executable path toward this goal.
RT-1 shows that scaling robot demonstration data improves language-conditioned policies~\cite{brohan2023}.
RT-2 transfers vision-language pretraining knowledge to action prediction~\cite{zitkovich2023}.
Open X-Embodiment expands task and embodiment diversity by aggregating heterogeneous robot data~\cite{openx2024}.
OpenVLA provides an open framework for VLA training and downstream adaptation~\cite{kim2025openvla}.
For continuous action generation, $\pi_0$ combines a pretrained vision-language backbone with a flow-matching action expert~\cite{black2025pi0}.
$\pi_{0.5}$ extends the training data mixture and generalization capability~\cite{black2025pi05}.
GR00T~N1 and the LingBot-VLA series explore foundation-scale robot policies across tasks and embodiments~\cite{bjorck2025groot,wu2026lingbot,wu2026lingbot2}.
These advances support adaptation to task-specific demonstrations.
\method builds on this foundation by introducing ordered task-progress representations into action-chunk generation.

\subsection{Temporal Planning for Robot Control}

Temporal modeling for robot control spans language abstraction, progress estimation, and latent-state representations.
RT-H introduces intermediate language-action descriptions between task instructions and low-level control~\cite{belkhale2024rth}.
BehaviorVLA learns temporally coherent behavior representations and incorporates phase information into action decoding~\cite{hu2026behaviorvla}.
ProgressVLA uses task-progress estimation to guide action prediction~\cite{yan2026progressvla}.
S$^2$-VLA maintains progress-related belief states for adaptive multimodal fusion~\cite{xie2026s2vla}.
These studies highlight the importance of explicitly representing task progress in long-horizon robot control.

Recent work further examines the correspondence between phase information and future actions.
Dense LSS aligns action tokens with reasoning representations of manipulation phases~\cite{li2026lss}.
ForeTime-VLA uses a world-action-model teacher and historical observations to transfer future-informative features into policy representations~\cite{ma2026foretime}.
GuidedVLA aggregates action-query features to predict a future skill distribution over a prediction interval~\cite{jia2026guidedvla}.
This distribution describes the proportions of phases within an action chunk.
\method further specifies phase order and transition positions through a separate prediction for each future action position.
The resulting plan features guide the corresponding action tokens.

%% file: sec/2_method.tex
\section{Method}
\label{sec:method}

\subsection{Overview and Problem Formulation}
\label{sec:overview}
\label{sec:problem}

\begin{figure*}[t]
  \centering
  \includegraphics[width=0.7\textwidth]{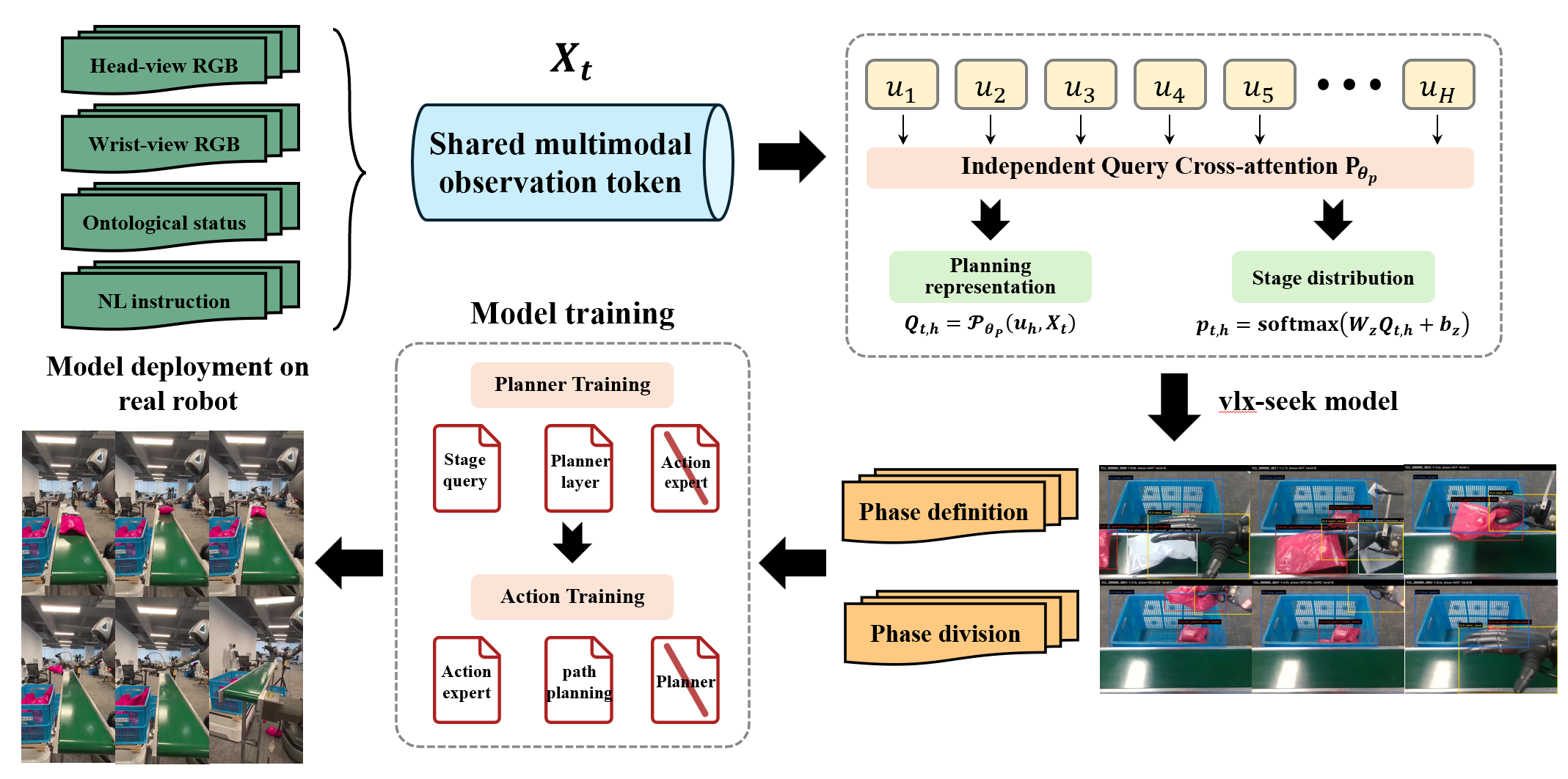}
  \caption{Overview of the \method framework.}
  \label{fig:phaseplan_framework}
\end{figure*}

Fig.~\ref{fig:phaseplan_framework} presents the overall framework of \method.
An ordered future-phase plan links multimodal observations to action-chunk generation.
Each future position has a phase representation that guides the corresponding action.
The framework combines position-aligned planning, plan-conditioned action generation, and two-stage adaptation within a unified policy.
The detailed quantitative evaluation uses the pretrained $\pi_{0.5}$ implementation.

At time $t$, the policy receives head-view RGB, wrist-view RGB, robot state, and a language instruction:
\begin{equation}
  o_t=(I_t^{\mathrm{head}},I_t^{\mathrm{wrist}},s_t,\ell).
  \label{eq:observation}
\end{equation}
The head view captures global spatial relations, while the wrist view provides local manipulation details.
The robot state describes the current configuration, and the instruction specifies the manipulation goal.
The policy predicts a future action chunk of length $H$:
\begin{equation}
  A_t=(a_{t+1},\ldots,a_{t+H}),\qquad a_{t+h}\in\R^{d_a}.
  \label{eq:action_chunk}
\end{equation}
The conveyor experiments use $H=16$.
During training, each future action position has a task-phase label $z_{t+h}\in\mathcal{Z}_{\tau}$.
Here $\mathcal{Z}_{\tau}$ denotes the phase vocabulary of task $\tau$.
The conveyor task contains four phases: \textsc{Wait}, \textsc{Act}, \textsc{Release}, and \textsc{Return-Home}.
Other manipulation tasks can define phase vocabularies that fit their own structure.
For the conveyor demonstrations, phase labels are constructed from per-frame VLX-Seek object annotations, recorded gripper state, and task-specific temporal rules.
These labels provide position-wise supervision over each valid future action window.
The phase-definition and phase-division components in Fig.~\ref{fig:phaseplan_framework} illustrate this process.

\method uses these labels to learn an internal ordered future plan $\hat Z_t=(p_{t,1},\ldots,p_{t,H})$.
Here $p_{t,h}$ is the predicted phase distribution for the $h$-th action position.
At deployment, the policy internally generates phase distributions of the same form from the standard observation $o_t$.

\subsection{Ordered Future-Phase Planner}
\label{sec:planner}

Let $X_t=E(o_t)$ denote the multimodal observation tokens obtained before action generation.
We introduce $H$ learnable phase queries $u_1,\ldots,u_H$, each corresponding to one future action position.
Every query reads the same observation context while retaining its own future-position identity:
\begin{equation}
  Q_{t,h}=\mathcal{P}_{\theta_P}(u_h,X_t),\qquad
  p_{t,h}=\operatorname{softmax}(W_zQ_{t,h}+b_z).
  \label{eq:ordered_planner}
\end{equation}
Here $Q_{t,h}$ is a contextual planning representation.
The distribution $p_{t,h}$ estimates the phase associated with action $a_{t+h}$.
The final planner uses mutually independent future queries.
The shared observation context describes the current scene, while different query embeddings specialize prediction to individual action offsets.
The ordered structure arises from the one-to-one correspondence $u_h\leftrightarrow z_{t+h}\leftrightarrow a_{t+h}$.

For the set of valid future positions $\mathcal{V}_t$, the planner is trained with a per-position phase classification loss:
\begin{equation}
  \mathcal{L}_{\mathrm{phase}}
  =-\frac{1}{\sum_t|\mathcal{V}_t|}
    \sum_t\sum_{h\in\mathcal{V}_t}
    \log p_{t,h}[z_{t+h}].
  \label{eq:phase_loss}
\end{equation}
This per-position objective preserves both the phase categories and the temporal locations of phase transitions inside the action chunk.

\subsection{Plan-Conditioned Action Generation}
\label{sec:plan_conditioning}

At each future position, the planner provides a contextual representation $Q_{t,h}$ and a phase distribution $p_{t,h}$.
A learnable matrix $E_z$ maps the distribution to a soft phase embedding.
We fuse this embedding with the planning representation:
\begin{equation}
  \bar e_{t,h}=E_z^{\top}p_{t,h},\qquad
  e_{t,h}=W_QQ_{t,h}+W_P\bar e_{t,h}.
  \label{eq:plan_feature}
\end{equation}
The feature $e_{t,h}$ contains both task-context information and explicit phase probabilities.
Let $x_{t,h}$ be the action-expert representation at future position $h$.
A lightweight network predicts the scaling and shift parameters for this position:
\begin{equation}
  (\gamma_{t,h},\beta_{t,h})=f_{\theta_M}(e_{t,h}),
  \label{eq:modulation_params}
\end{equation}
and modulates the corresponding action token:
\begin{equation}
  \tilde x_{t,h}
  =(1+\gamma_{t,h})\odot\operatorname{RMSNorm}(x_{t,h})
   +\beta_{t,h}.
  \label{eq:position_conditioning}
\end{equation}
The final projection of $f_{\theta_M}$ is zero-initialized.
The modulation path therefore preserves the behavior of the pretrained action path at the beginning of adaptation.
Each $e_{t,h}$ affects only the corresponding $x_{t,h}$.
This maintains position alignment between the plan and the action across the whole action chunk.

\subsection{Training and Inference}
\label{sec:two_stage}

\method separates future-plan learning from action adaptation.
In the first stage, $\mathcal{L}_{\mathrm{phase}}$ is optimized to train the phase queries and planner layers.
The goal is to predict an ordered future phase sequence from the current observation.
The action expert is held fixed during this stage.
The selected planner is then frozen.
The second stage optimizes the original flow-matching action objective $\mathcal{L}_{\mathrm{flow}}$.
It updates the action expert and the per-position modulation path.
Throughout action adaptation, the frozen planner continues to provide $(Q_{t,h},p_{t,h})$.

This decomposition gives the planning representations stable temporal semantics before they guide action generation.
The complete inference computation is
\begin{equation}
  o_t\longrightarrow(Q_{t,1:H},p_{t,1:H})\longrightarrow A_t.
  \label{eq:inference}
\end{equation}
The planner and the conditioning path reside in the same \method model.
The deployment inputs follow the base VLA policy: RGB images, robot state, and language instruction.
The model generates its phase plan internally from the current observation and uses the resulting features to guide future actions.
Object annotations and constructed phase labels serve as training supervision.
The real-robot sequence in Fig.~\ref{fig:phaseplan_framework} illustrates deployment of the complete policy on the conveyor task.

%% file: sec/3_experiments.tex
\section{Experiments}
\label{sec:experiments}

\subsection{Task and Experimental Setup}
\label{sec:conveyor_dataset}

We evaluate \method on a dynamic conveyor task collected through real-robot teleoperation.
The robot observes head-view RGB, right-wrist RGB, proprioceptive state, and a language instruction.
Each policy prediction outputs an action chunk of length 16.
Each action step contains seven right-arm joint targets and one binary hand open or close command.

\begin{figure*}[t]
  \centering
  \includegraphics[width=0.7\textwidth]{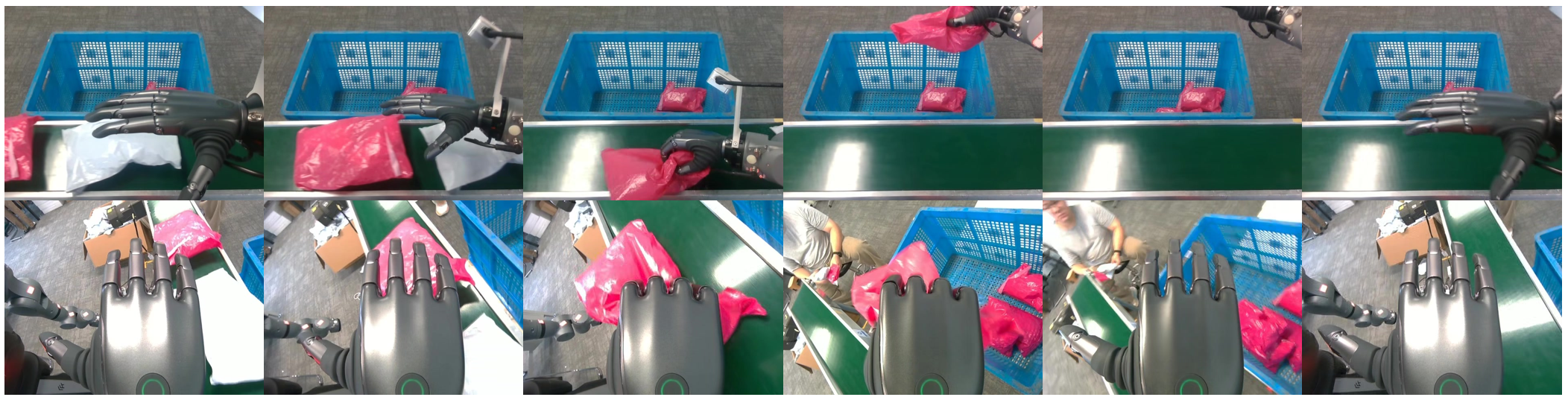}
  \caption{Dual-view observations from a training trajectory.}
  \label{fig:dual_view_observation}
\end{figure*}

Fig.~\ref{fig:dual_view_observation} shows examples of dual-view observations from a training trajectory.
Each column corresponds to synchronized images at the same time instant.
The top row is the head view, and the bottom row is the right-wrist view.
The head view provides the overall spatial relations among the conveyor, basket, packages, and robot arm.
The right-wrist view presents the local geometry between the hand and the package to be manipulated.
The two views are complementary in scene coverage and manipulation detail.
Together, they support inference of object states, task progress, and subsequent actions.

The conveyor contains actionable red packages and white packages that serve as distractors.
Each cycle starts with waiting for an actionable package.
The robot approaches and grasps it, then transports it to a blue basket.
After release, the arm returns to its initial pose.
We represent this process with four phases:
\begin{equation}
  \textsc{Wait}\rightarrow\textsc{Act}\rightarrow
  \textsc{Release}\rightarrow\textsc{Return-Home}.
  \label{eq:conveyor_phases}
\end{equation}

\begin{figure*}[t]
  \centering
  \includegraphics[width=0.7\textwidth]{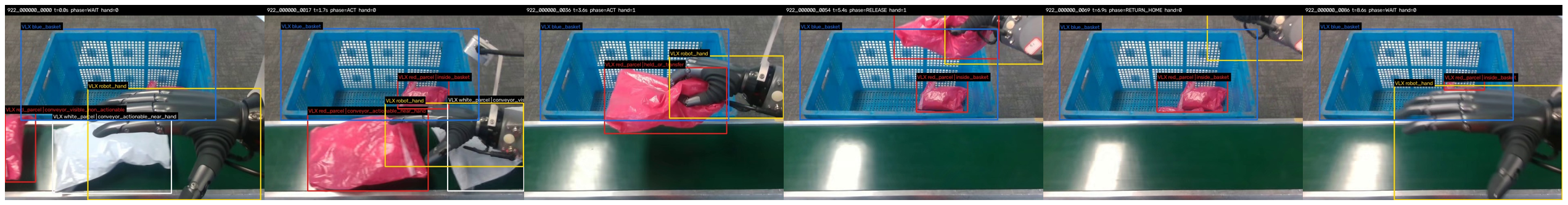}
  \caption{Object annotations and constructed task-phase labels.}
  \label{fig:vlx_seek_annotation}
\end{figure*}

Fig.~\ref{fig:vlx_seek_annotation} shows a sequence with VLX-Seek object annotations and constructed phase labels.
Each frame displays target-level bounding boxes alongside its task-phase label.
Blue boxes identify the basket, and red boxes identify the target package.
Yellow boxes mark the robot arm or gripper.
White boxes mark other visible packages or relevant objects on the conveyor.
The frame sequence covers waiting, approaching and grasping, releasing, and returning home.
It directly illustrates the temporal correspondence between visual object states and task progress.
We combine the object annotations with recorded gripper state and task-specific temporal rules to construct demonstration phase labels.
The resulting labels align phase transitions with the corresponding future action positions.

The dataset contains 325 trajectory episodes and 40,129 frames.
Before training, we fix the data split by episode.
Every valid 16-step action window has phase supervision labels aligned position by position with the future actions.

Data preprocessing removes teleoperation jitter in stationary \textsc{Wait} phases and smoothly re-anchors the \textsc{Wait}-to-\textsc{Act} transition.
All compared methods use the same processed action targets.
The comparison therefore focuses on the ordered planning and conditioning mechanisms.

\subsection{Baselines and Metrics}
\label{sec:baselines}

We compare \method with three variants of the same backbone.
All methods share the $\pi_{0.5}$ initialization, demonstration data, episode split, optimization budget, and action representation.
\textbf{Original $\pi_{0.5}$} directly adapts the original action expert module.
\textbf{Future-skill histogram} follows the temporal-summary formulation used in GuidedVLA~\cite{jia2026guidedvla}.
We construct this baseline using an aggregated distribution of future skills.
\textbf{Ours-P} trains the ordered future-phase planner alongside the original action path.
This separates phase prediction itself from its effect on action generation.
\textbf{\method} further connects the plan feature of each position to the corresponding future action token.

The planner ablations compare independent queries with interactive queries.
They also test an input that contains only state and language, and an input with shuffled RGB images.
The action-side ablations examine conditioning with the phase ID only.
They further test continued planner optimization during action adaptation and an oracle upper bound that uses ground-truth future phase labels.
All variants use the same evaluation windows and random seeds.
The reported metrics include future-phase prediction, phase-transition quality, joint-action error, hand-command accuracy, and performance on cross-phase action chunks.

%% file: sec/4_results.tex
\section{Results}
\label{sec:results}

\subsection{Future-Phase Planning and Action Prediction}
\label{sec:planning_results}

On the conveyor validation set, the independent-query planner achieves about 92.5\% per-position future-phase accuracy.
Its phase-transition-type accuracy is about 89\%.
Prediction accuracy decreases gently as the future prediction horizon increases.
It remains high across all 16 positions.

The visual-input ablation confirms the contribution of current images to phase prediction.
With only robot state and language, the macro-averaged F1 is about 43\%.
Shuffling the RGB images reduces macro-averaged F1 further to about 36\%.
The full vision-language-state planner exceeds 91\%.
On the main phase metrics, independent and interactive queries perform closely, with independent queries slightly better.
The final model therefore adopts the simpler independent-query structure.

\label{sec:action_results}

\begin{table}[t]
  \centering
  \caption{Offline action prediction on the conveyor validation set.}
  \label{tab:action_results}
  \footnotesize
  \setlength{\tabcolsep}{2.5pt}
  \begin{tabular}{lccc}
    \toprule
    Method & \makecell{Overall joint\\MAE $\downarrow$} & \makecell{Cross-phase joint\\MAE $\downarrow$} & \makecell{Hand\\accuracy $\uparrow$} \\
    \midrule
    Original $\pi_{0.5}$ & 0.044653 & 0.051292 & 90.394\% \\
    \makecell[l]{Future-skill\\histogram} & 0.039619 & 0.046721 & 91.864\% \\
    Ours-P & 0.044653 & 0.051292 & 90.394\% \\
    \textbf{\method} & \textbf{0.034605} & \textbf{0.041318} & \textbf{93.162\%} \\
    \bottomrule
  \end{tabular}
\end{table}

Table~\ref{tab:action_results} summarizes the offline action-prediction results.
Results are averaged over three random seeds.
The validation set contains 3,584 windows, including 1,613 that cross phase boundaries.
Joint-action MAE is measured in radians; lower values indicate better performance.
Higher hand-command accuracy indicates better performance.
\method achieves the lowest joint-action error on both the full validation set and the cross-phase action chunks.
Its overall joint MAE is 0.034605\,rad.
This is 22.5\% lower than original $\pi_{0.5}$ and 12.7\% lower than the future-skill-histogram baseline.
On cross-phase windows, the MAE drops to 0.041318\,rad, and the hand-command accuracy reaches 93.162\%.

Ours-P learns an ordered future-phase sequence and retains the action predictions of the original $\pi_{0.5}$.
The comparison between Ours-P and \method therefore isolates the effect of per-position plan conditioning.
Connecting each plan representation to its corresponding action position enables the predicted plan to improve action generation.

\subsection{Simulation and Real-Robot Validation}
\label{sec:mujoco_validation}

We further examine the executability of predicted actions in MuJoCo using recorded visual observations and continuously updated simulated robot states.
The evaluation uses samples from the validation set.
The head-view and right-wrist RGB images come from the original validation-set sequences.
The robot joint and hand states are provided in real time by the simulated robot in MuJoCo.
At each decision step, the policy receives these visual, state, and language inputs.
It predicts a 16-step action chunk, which is executed by the simulated robot.
We examine the resulting evolution of the robot pose.

\begin{figure*}[t]
  \centering
  \includegraphics[width=0.7\textwidth]{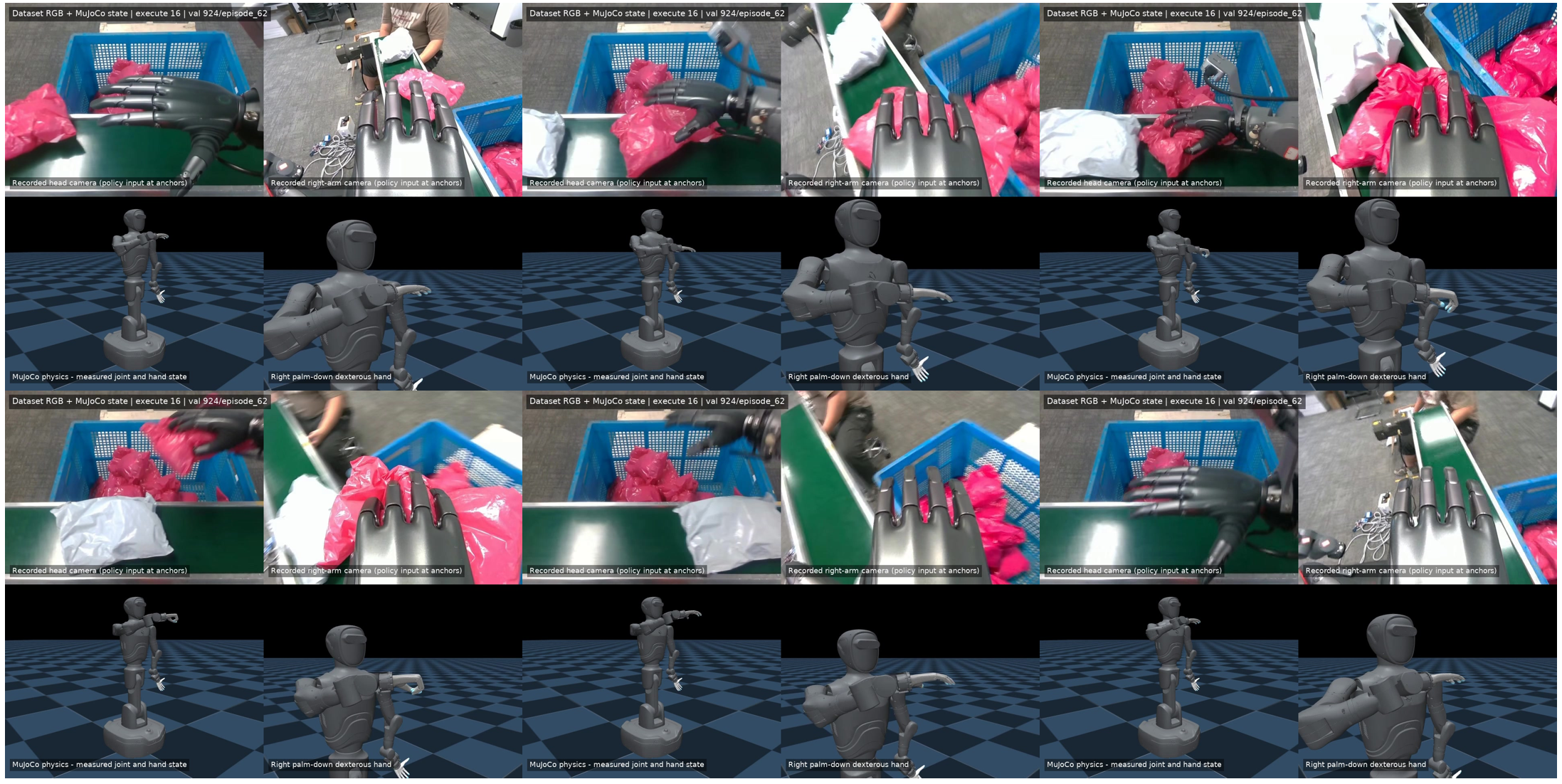}
  \caption{MuJoCo action execution with recorded visual inputs.}
  \label{fig:mujoco_validation}
\end{figure*}

Fig.~\ref{fig:mujoco_validation} shows several representative instants from a validation sequence.
At each instant, the upper part contains the original dual-view images fed into the policy.
The lower part shows the full-body and right-hand states after executing the predicted joint and hand commands.
The simulated robot produces continuous arm-pose and hand-configuration changes along the input sequence.
This indicates that the predicted actions map to continuous robot motion in physical simulation.
This hybrid evaluation provides a controlled view of how predicted commands evolve under robot kinematics and simulated physical constraints.

\label{sec:real_robot_validation}

\begin{figure}[t]
  \centering
  \includegraphics[width=0.7\linewidth]{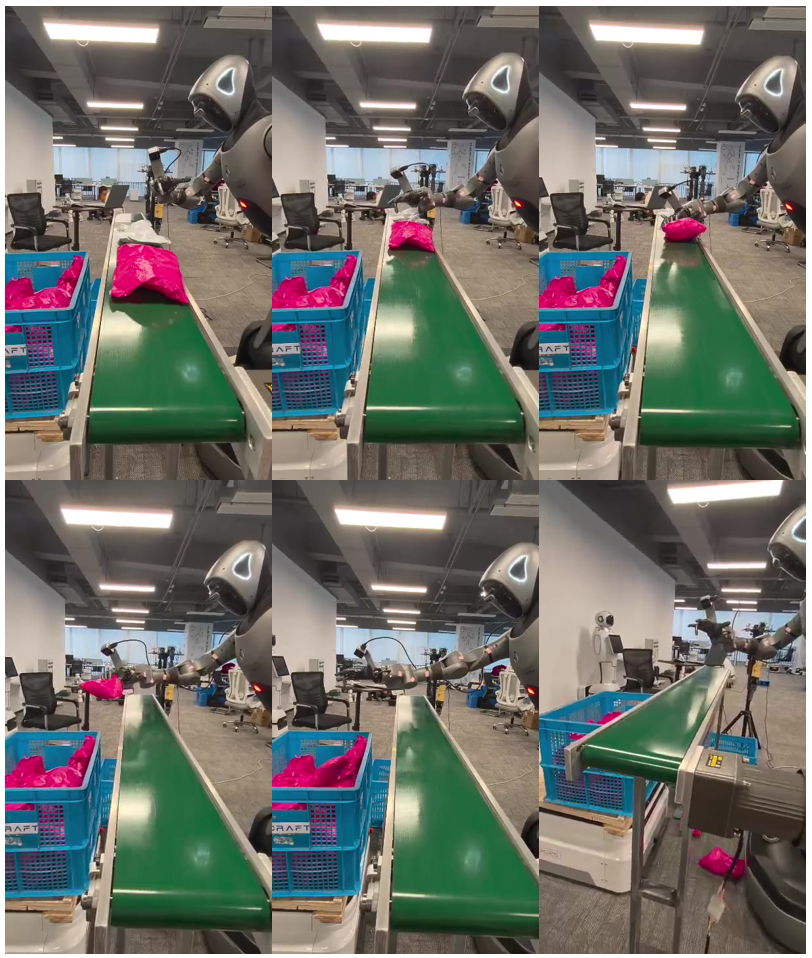}
  \caption{Real-robot deployment on the conveyor task.}
  \label{fig:real_robot_validation}
\end{figure}

We further deploy \method on the physical conveyor robot to examine its behavior under real visual feedback and robot dynamics.
The real-robot test uses the same visual observations, robot state, and language instruction as the training task.
The policy generates action chunks online and drives the robot arm.
Fig.~\ref{fig:real_robot_validation} shows representative frames from a real-robot manipulation sequence.
The robot approaches and grasps a red package from the conveyor.
It then transports the package to the blue basket and completes placement.
Across consecutive frames, the package position, arm pose, and gripper state change coherently with task progress.
The sequence illustrates continuous execution from grasping to placement under real visual feedback and robot dynamics.
This qualitative example complements the offline metrics and MuJoCo results by illustrating behavior on the physical system.

\begin{figure*}[t]
  \centering
  \includegraphics[width=0.7\textwidth]{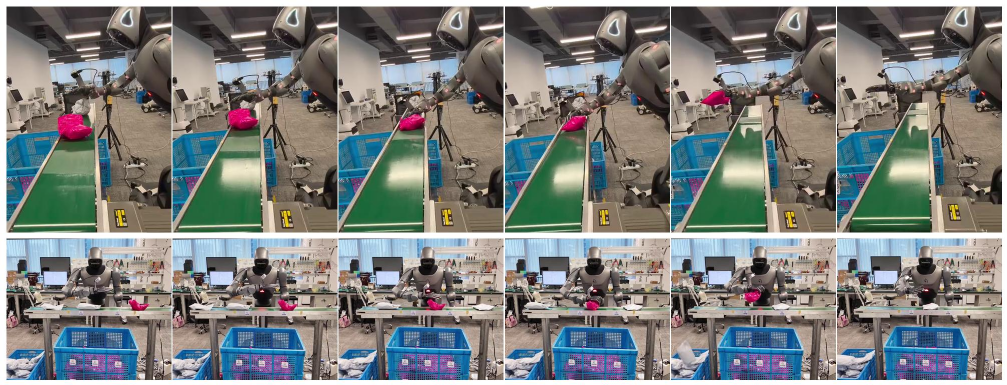}
  \caption{Real-robot execution on AcrossWAM1.0.}
  \label{fig:acrosswam_real_robot}
\end{figure*}

To examine the strategy beyond its $\pi_{0.5}$ instantiation, we also deploy it on AcrossWAM1.0~\cite{zhang2026acrosswam10}.
Fig.~\ref{fig:acrosswam_real_robot} shows qualitative real-robot executions from side and frontal viewpoints.
The visible progression from approach and grasp to transport and release provides qualitative evidence that our strategy is also effective on AcrossWAM1.0.
The top row presents an oblique side view, and the bottom row a frontal view; the rows show separate conveyor-sorting sequences.
In each sequence, the robot approaches a pink package on the green conveyor, grasps it, moves it toward the blue collection basket, and ends with an empty gripper.
The ordered frames illustrate coherent task progression across both viewpoints.

\subsection{Ablation Studies}
\label{sec:ablation_results}

Full plan features yield lower action error than phase-ID conditioning.
This shows that the contextual planning representation $Q_{t,h}$ provides useful information beyond the phase distribution $p_{t,h}$.
It also supports the fusion of both signals in the per-position feature $e_{t,h}$.

Two-stage training balances planning quality and action adaptation.
Continuing to update the planner during action training slightly improves the average per-position phase accuracy.
Freezing the planner improves phase-transition-type accuracy and transition timing.
It also yields slightly lower joint-action error.
The final model therefore adopts the frozen planner.
A ground-truth future-phase oracle further improves action prediction.
The remaining gap is largest for \textsc{Release}-related hand predictions.
This result highlights the importance of accurate transition prediction for short manipulation phases.

\label{sec:design_studies}

Explicit box-level visual supervision raises the proportion of attention falling inside annotated regions from about 8.5\% to 81.2\%.
The corresponding action metrics are comparable.
The final architecture adopts the simpler ordered plan-conditioning path.

Inverse-phase-frequency weighting improves hand-command accuracy at future \textsc{Release} positions from about 82.6\% to 84.3\%.
Overall joint-action MAE increases by about 1.3\%.
Cross-phase action error also increases slightly.
The uniformly weighted flow-matching objective provides a better overall trade-off and is used in the final model.

%% file: sec/5_conclusion.tex
\section{Conclusion}
\label{sec:conclusion}

We presented \method, an ordered future-phase planning framework for robot brain models.
The framework represents task progression within an action chunk through a phase distribution for each future action position.
Each contextual planning representation and phase distribution jointly condition the corresponding action token.
This mechanism preserves phase order and transition locations throughout the prediction horizon.
A two-stage training strategy first learns future-phase representations, then freezes the planner during action adaptation.
The action model thus learns position-aligned control from plans with stable temporal semantics.
Planning and action generation remain integrated within a single policy using standard multimodal inputs.

Conveyor-belt experiments demonstrate the effectiveness of this design in the $\pi_{0.5}$ implementation.
The planner achieves about 92.5\% per-position phase accuracy and 89\% phase-transition-type accuracy.
Plan-conditioned action generation reduces offline joint-action error by 22.5\% relative to the original policy and 12.7\% relative to the future-skill-histogram baseline.
The gains also extend to action chunks that cross phase boundaries.
Ablations support the use of contextual plan features and a frozen planner for accurate action prediction and transition timing.
MuJoCo execution with recorded visual inputs produces continuous robot motion as simulated states evolve.
Physical deployment further illustrates coherent execution from package grasping to placement under real visual feedback.
Qualitative deployment on AcrossWAM1.0 also shows coherent task progression from approach and grasp to transport and release.
Together, these results establish ordered future-phase planning as a practical way to incorporate task progress into robot brain models.
Position-aligned phase representations connect task-level temporal structure with continuous action generation.
